\pdfoutput=1
\documentclass[11pt]{article}

\usepackage{amsmath, amsthm, amssymb}
\usepackage{graphicx} 
\usepackage{wrapfig}
\usepackage{hyperref}
\begin{document}
\title{The Convergence of AI code and Cortical Functioning -- a Commentary}
\author{David Mumford}
\date{Emeritus Professor of Applied Mathematics, Brown University\\[1em]
\today}
\maketitle

\begin{abstract} Neural nets, one of the oldest architectures for AI programming, are loosely based on biological neurons and their properties. Recent work on language applications has made the AI code closer to biological reality in several ways. This commentary examines this convergence and, in light of what is known of neocortical structure, addresses the question of whether ``general AI'' looks attainable with these tools.
\end{abstract}

\noindent One of the earliest ideas for programming Artificial Intelligence was to imitate neurons and their connectivity with {\it neural nets}. In the turbulent boom and bust evolution of AI, this remained a theme with strong adherents, but it fell out of the mainstream until around 2010 when these ideas were implemented with really huge datasets and really fast computers. The field of AI has now had a decade of tremendous progress in which neural nets, along with some major improvements, have been the central character. The purpose of this post is to describe the further parallels between the software implementation of AI and the instantiation of cognitive intelligence in mammalian brains. I conjecture that, for better or for worse, all future instances of artificial intelligence will be driven to use these algorithms even though they are opaque and resist simple explanations of why they do what they do. This commentary follows my blog post \href{http://www.dam.brown.edu/people/mumford/blog.html}{here}.

\section{Neural Nets}

Rectifying neural nets (ReLU nets), mathematically speaking, are just a class of piecewise linear functions $f:\mathbb{R}^n \rightarrow \mathbb{R}^m $ defined in a very specific way, as a composition of simple factors, all given by formulas of the following type:
$$ \phi(\vec x) _i = \max\left[ 
0, \left( {\textstyle \sum_j M_{ij}*x_j }+ b_i\right) 
\right] $$
Such a function is always diagrammed as a set of layers, with functions \( \phi \) computing the next higher layer from the layer below and the running value after each composition being called the ``activity'' $\vec x^{(n)}$ in layer $n$. The components of these activities are called ``units'', as these are supposed to correspond to neurons in the biological interpretation. The whole net depends on the weight matrices $M^{(n)}$ and the bias vectors $\vec b^{(n)}$ for each level, all of which need to be learned by fitting data via gradient descent, called ``back propagation'' from the shape of the formula for the gradient.  It is easy to see that $f$ is a piecewise linear continuous function: if the vector space of inputs is divided into polyhedral cells, each defined by the set of units whose activity is zero, then the output is a linear function on each of these cells. This whole apparatus is just an example of regressing data with a particular class of functions. 

A miracle (for which to my knowledge nobody yet has a good explanation) is how well gradient descent works to train the neural net: tested on new data, its performance is usually not much worse than on the training data. Somehow, it rarely overfits the training data even if it has a truly huge number of weights.  Except for some small bells and whistles, this is the whole thing.

The motivation for this algorithm was an extremely simplified model of what animal neurons do. Neurons in all animals, from a jelly fish up, do form a net in which every neuron has a single axon that branches and contacts the dendrites of other neurons at synapses. Electrical signals do indeed propagate from neuron to neuron. The signals, however, (with a few exceptions) come in short (1 or 2 milliseconds) pulses, its spikes, so the message sent from one neuron to another is called a spike train. Simplification \#1:  take the rate of firing, spikes per second, as a real number signal emitted by each neuron. Thus, in neural nets, each unit is taken to correspond to one neuron, its real value $x^{(n)}_i$ being the associated firing rate. And when does the receiving neuron emit a spike in its turn? Simplification \#2:  assume that all neurons add up their active dendritic inputs linearly with weights indicating the strength of each synaptic connection (positive if it is an excitatory stimulus to the receiving neuron, negative if it is inhibitory) and that their firing rate is this sum after some kind of rectifying function is applied because the firing rate must be positive. Bingo: this is exactly what the function $\phi $ does, so we now have a neural net that is a rough caricature  of the biological reality. Well, there is also Simplification \#3: assume that neural synapses do not form loops so we can put the neurons in layers, each speaking only to neurons above them. {\it Unfortunately, none of these simplifications are true.}

Some modifications that make the neural nets a bit more realistic have been known for some time to also make them work better. First, there is no rigid layer structure in the cortex and neural nets often work better when there are layer skipping links, i.e. layer $n$ can have some inputs from layer $n-2$ or lower. A special case are ``residual" networks where the variable $\vec x^{(n-2)}$ is added to the variable $\vec x^ {(n)}$, forcing the intermediate layers to seek not a totally new signal but a correction to $\vec x^{(n-2)}$. Another modification, known as ``dropout", trains the network to work even with a certain percent of variables $x^{(n)}_ i$ set to zero. This forces the neural net to be redundant just as our thinking seems to be resilient to some neurons malfunctioning. A third improvement is called ``block normalization''. This introduces an extra variable at each unit  that, together with its bias, moderates the mean and variance of each unit's response to a random batch of data, something like regulating chemicals in the neuron.

\section{Tokens vs.\ distributed data}

The neural recordings of David Hubel and Torsten Wiesel in the 1960's found a remarkable thing. They recorded from V1, the primary visual cortex, in cats, and discovered that each neuron seemed to have a definite preferred stimulus, like a bar or blob in a specific location on the cat's retina, that caused the neuron to fire. This led to Simplification \#4, that {\it all} neurons were waiting for some event, some situation, stimulus or planned movement and that they fired in its presence. This was called the {\it grandmother hypothesis}, e.g.\ there should be a cell somewhere in the brain that fires if and only if you are looking at your grandmother. More to the point, for each word we hear or speak, there should be a cell which fires when that word is heard or pronounced. If the grandmother hypothesis were true, all we needed to do was figure out the dictionary, neuron to stimulating situation, and an exhaustive recording of neural activity would tell us what the animal is ``thinking". Although there were a few successes in this direction, it hit a brick wall when recordings were made in the higher visual area V4 and in the visual inferior temporal cortex (IT). It was quickly discovered that these cells were indeed paying attention to visual input and seemed to be looking at more complex features of the retinal signal: shapes, textures, perhaps the identity of objects in the scene. But no one could pin this down because there seemed an explosive number of combinations of features that stimulated each cell to varying degrees. In other words, the {\it simultaneous firing pattern} of large populations of cells seemed to carry the information, instead of each cell separately telling us one thing about the stimulus. Thus the stimulus seems to be encoded as a high dimensional vector that captures what was going on, perhaps thousand dimensional or more. The information is distributed over an area in the cortex and there is no simple meaning in single cell firings.  Here's a new confirmation of neural net architecture: the idea that it is the simultaneous real values of all units in a layer that carries the data while the values of single units have no easy interpretation.

Meanwhile, the AI people are trying to solve problems not only with understanding images but especially understanding language. Raw images are represented already by a big vector of real numbers, the values of their pixels. Typical problems are face recognition and general object recognition. Words, on the other hand, are just a list in a dictionary. Typical problems are sentence parsing, machine translation and internet question answering. How should neural nets be applied to such language tasks? A  breakthrough arrived when a team of researchers at Google published an algorithm in 2013 called {\it  word2vec}\footnote{arXiv: 1301.3781v3}. The idea was to represent each word as a real valued vector in a high dimensional vector space, an instance of what has been called {\it vector symbolic architecture}.  The constraint was that words which often occur near each other in speech or written text should correspond to nearby vectors, their distance reflecting how often they co-occur.  One way to think of this is that a word has many aspects to it such as its syntactic role, its semantic classification in many senses, as well as other reasons why it co-occurs with other words, and high dimensional vectors have enough freedom to be able to capture much of this. What is remarkable is that this represents a major convergence of AI programs with actual neural activity. Needless to say, no neurons have ever been found the human brain that respond to a single word and only that word\footnote{Recordings from the exposed brain of awake patients are employed in some operations for severe epilepsy}.

\section{Transformers and context}

Remarkably, the Google language team went further in the 2017 paper entitled {\it Attention is all you need}\footnote{arXiv: 1706.03762}.  It introduces a completely new architecture that enhances neural nets in a powerful way, the {\it transformer}. The authors are looking at linguistic tasks involving whole sentences which are taken in one word at a time and build word representations that encode the meaning of the word {\it in the context of a whole sentence}. The linguistic tasks they sought to solve all involve outputting a new sentence, e.g.\ translating the input sentence into a second language, answering the question posed by the first sentence or more simply filling in a word that was purposely omitted from the first sentence. Thus the algorithm has two parts: an encoder creating vector representations of all input words and a decoder reversing the process producing a new sentence.

What does the transformer do?  Looking at the encoder, for 6 layers of a conventional neural net, they add an extra layer each with 8 separate {\it attention heads}.  For each of these heads in each layer, one trains three linear maps from the 512 dimensional layer data to shorter 64 dimensional vectors, the maps being called {\it queries}, {\it key} and {\it values} respectively.  Remarkably, this requires training from data $6 \times 8 \times 3 \times 64 \times 512$ (or roughly 5 million) more coefficients that need to be trained!  The idea is first to find, for each query applied to the current word (by matrix multiplication with the vector for this word at this level), the key applied to some other word in the sentence that is closest to it: then scale a measure of the distance between this query and this key to [0,1]; and finally use this to weight and then add up the corresponding value vectors. Concatenate these over the 8 heads, bringing the dimension back up to 512 and train a final $512 \times 512$ matrix to jumble it all up like a fully connected layer of a neural net and add this to original layer vector. In formulas, this comes out fairly simply like this:
$$ \text{Cat}_h \sum_\alpha \text{softmax}_\alpha \left(C.(X.W_h^Q).(Y_\alpha .W_h^K)^t \right) Y_\alpha W_h^V$$
where ``Cat" stands for concatenate,  $h$ indexes the heads,``softmax" means exponentiating and dividing by the sum, $C$   is a constant, $X$  is the input vector, $Y_\alpha$ are context vectors, the $W^Q$'s are the query matrices, $W^K$ the keys and $W^V$ the values. 

One of the most convincing demonstrations of what transformers do comes from the 2019 paper {\it A Structural Probe for Finding Syntax in Word Representations} by Chris Manning and John Hewitt\footnote{arXiv: 1803.00188}. They took the public domain Google program ``BERT-large'' that, when given a database of sentences, produces vector representations of all its words. The program comes with fixed queries, keys and values from its training on two tasks: i) predicting words randomly excised from normal English sentences via its "decoder" and ii) the task of determining, for a pair of sentences, whether the second was a logical continuation of the first or has nothing to do with it. The point here is that it has {\it not}   been trained on any tasks involving syntax. They then took sentences with known parse trees and looked for low dimensional projections of BERT's word representation at various levels such that, for any two words in the sentence, the squared distance between their projected word vectors approximated how many links in the parse tree connected the two words. Amazingly, they found that the best projections to say 20 dimensions allowed them to predict the true parse tree with 80\% accuracy. In other words, the transformers are finding the underlying syntax of the sentence, but hiding it in the vector representation, and using it in order to solve the missing word or the sentence continuation problem. This goes a long way, I think, to clarifying why these programs are so good at language translation.

The really significant conclusion of this demonstration is that, yes -- the neural net is learning syntax, but no -- it doesn't make {\it explicit} use of the syntax to solve problems. A number of researchers, including myself with Song-chun Zhu\footnote{A Stochastic Grammar of Images, {\it Foundations and Trends in Computer Graphics and Vision}, {\bf 2,4} (2006); also CVPR workshop, 2009, \href{http://www.dam.brown.edu/people/mumford/vision/papers/2009--GeneralGrammars.pdf}{\bf (link)} }, have argued that versions of the parse tree of sentences pervade every kind of thinking, so that some sort of graphical representation is needed to explain all cognitive thinking. This is a fundamental principle in Grenander's {\it Pattern Theory} and has been widely implemented by Song-chun Zhu's AND/OR graphs. It appears, however, that these graphs need not be an explicit part of cognitive algorithms, that they may merely be implicit. I will return to this in a discussion of the Whorfian hypothesis below.

\section{Context in the brain}
The take away from the success of transformers would seem to be that calculations that incorporate context require  more than the simple weighted summation of vanilla neural nets. And, indeed, it has also long been clear that {\it neurons do a great deal more than add up their synaptic inputs}. To explain this, we need to review more of the basic biology. 

\subsection{Pyramidal cells}
The cortex is the structure common to all mammals which clearly is responsible for their cognitive intelligence (as opposed to muscular skills, instinctive emotional responses and routine behaviors). It is composed of six layers, each with its distinctive neurons and connections. Something like 2/3rds of its neurons are {\it pyramidal cells}, large excitatory neurons oriented perpendicular to the cortical surface, with up to 30,000 synapses in humans.  They are the workhorses of cognition. They occur in most layers, as shown in the figure below\footnote{From G.Radnikow and D.Feldmeyer, ``Layer- and Cell Type-Specific Modulation of Excitatory Neural Activity in the Neocortex'', {\it Frontiers in Neuroanatomy}, 2018}.\\
\begin{figure}[h]
\centering
\includegraphics[width=5in]{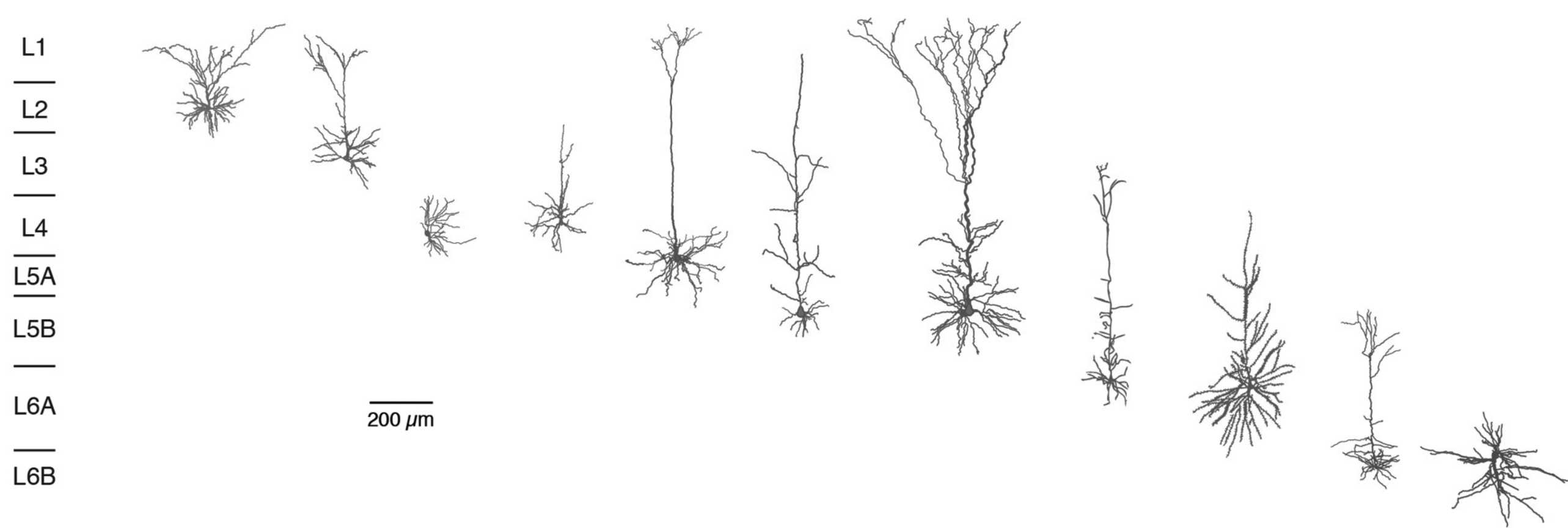}
\caption{\small Sample dendritic arbors of excitatory cells from mouse cortex with cortical layers shown on the left. All are pyramidal except for the 3rd (a {\it stellate} L4 cell) and last (a {\it multipolar} L6B cell). The cell body, called the {\it soma}, is the dark blob near the bottom of each cell. All the lines are dendrites, those at the top called ``apical'', those at the bottom ``basal''}
\end{figure}

Modelers have long known that a pyramidal cell does something more complex than simply add up these 30,000 inputs.  For one thing, their dendrites are not merely passive conductors but they have voltage gated channels that, like the axon, allow them to create moving spikes\footnote{See N. Spruston, G. Stuart and M. Hausser, Principles of dendritic integration, in {\it Dendrites}, Oxford Univ. Press, 3rd edition, 2016}. These can propagate either from synapses to the soma or, retroactively, from soma to synapses. In addition, they have special receptors on their basal dendrites, the {\it NMDA receptors}, that detect coincidence between arrival of new excitation and prior depolarization of the same part of the dendrite. These can depolarize part of the dendrite for periods of 100 milliseconds or more, known as NMDA plateaus or spikes\footnote{See Srdjan Antic et al, The Decade of the Dendritic NMDA Spike,  {\it J. Neuroscience Research}, {\bf 88}, 2010}.

One hypothesis, the ``Two Layer Model'', is that the various branches of its dendritic tree are each doing some first stage of a computation and then, in a second stage, the cell as a whole combines these in some fashion\footnote{See Bartlett Mel, Toward a simplified model of an active dendritic tree, in {\it Dendrites}, op.cit.}. But there is no consensus model for this yet, only suggestive bits and pieces. Another hypothesis is that, at any given time, some branches of the tree may be activated in such a way that its depolarization creates spikes in the dendrite that carry their responses to the soma, while other branches are silenced. This amounts to a set of gates on the branches, allowing the cell to compute quite different things depending on which branches are activated. Finally, when the cell fires, emitting a spike on its axon, it can also generate a back propagating spike in the dendrites, altering their subsequent activity perhaps in some context specific way.

It is tempting to seek a transformer-like algorithm that uses all this machinery. However we need to face one way in which the mechanisms of computers and brains will never converge: signals in the brain are trains of spikes, not real numbers. It is true that the membrane potential of a neuron is a real number (in fact, a real-valued function along each dendrite and axon) but the cell's output is a stereotyped spike, always identical. What varies between spike trains is the {\it timing} of the individual spikes. Brains have no central clock and many modelers have speculated that precise spike timings, especially synchronous spikes, are integral parts of the ongoing cortical computation. This could allow spike trains to carry much more information than merely its spike count.

The essential idea of a silicon transformer is to seek ways in which the signal $\vec x$ being analyzed has certain definite connections to some part $\vec y$ of the context (e.g.\ some other sensory data or some memory, etc.). Transformers do this by computing products $x\cdot M \cdot y$ for learned low rank matrices $M$. It's quite conceivable that interlaced synapses, some with NMDA receptors, along basal dendrites of pyramidal cells, could do something similar if they carry synapses for the both the $x$ and the $y$ signals. For example, see Bartlett Mel's paper cited above. The interaction of NMDA receptors versus the conventional (AMPA) receptors may well implement a nonlinear version of  $\sum_i x_i y_i $ . This might be the basis of a transformer-like mechanism linking {\it local} neurons, e.g.\ linking the signals from two words in a heard sentence or from two objects in a scene being viewed.

\subsection{Feedback}
However, there is another challenge about which I made speculations 30 years ago in the journal Biological Cybernetics \href{http://www.dam.brown.edu/people/mumford/vision/papers/1991a--BiolCyb-journal.pdf}{\bf (link)} and in an AMS Symposium  for Norbert Wiener
\href{http://www.dam.brown.edu/people/mumford/vision/papers/1997d--ModelingCortexWiener-NC.pdf}{\bf (link)}. It's well established in neuroanatomy that the cortex can be divided into high level and low level areas with processing streams going both ``forward'', e.g. from the primary sensory areas to association areas as well as ``backwards'', usually called feedback pathways. These connections are set up by long distance pyramidal axons in specific cortical layers and these have been meticulously worked out. A current diagram of these pathways\footnote{Anatomy of Hierarchy, Nikola Markov et al, {\it Journal of Comparative Neurology},  {\bf 522}, 2014} is reproduced below. 

\begin{figure}[h]
\centering
\includegraphics[width=5in]{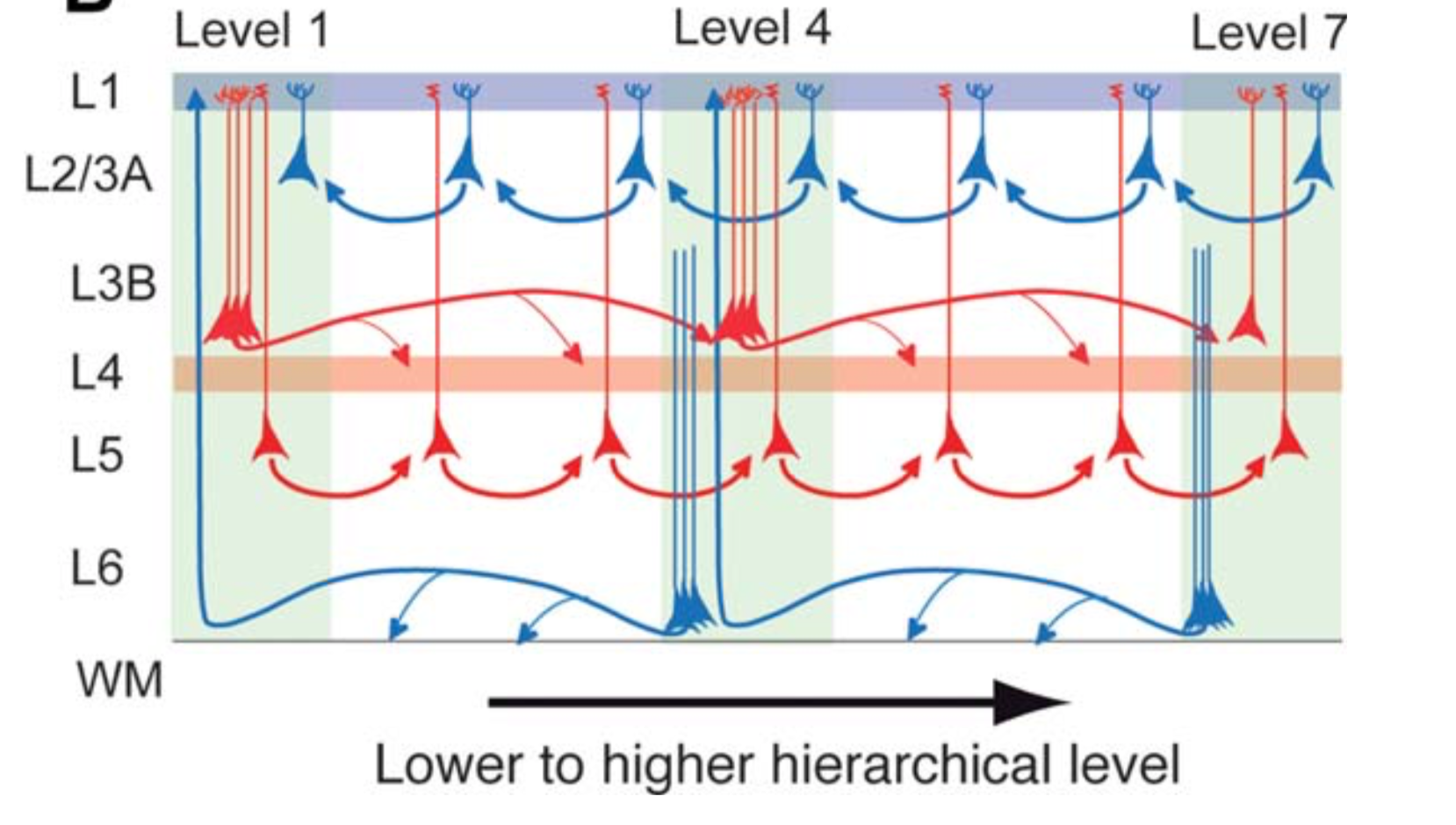}
\caption{\small The red arrows are feedforward processing involving layers 3B, 4 and 5, while the blue arrows are feedback pathways involving layers 1, 2, 3A and 6. WM is white matter. The triangles are pyramidal cell bodies with the vertical lines indicating their dendrites.}
\end{figure}

My proposal some decades ago was that feedback was connected computationally to Bayes's rule. Naively, the rule by itself could be implemented, for example, if the feedback path carried a vector of prior probabilities of possible high level states that was combined with the locally computed conditional probabilities by a dot product. My proposal was more complicated but whatever high level data is sent to a lower area, this is a natural place for biological versions of transformers. More specifically, I sought an architecture for connecting long term memories like knowledge of the sounds of words or of the shape of objects, etc.\ to current sensory data but registering their differences. For all such tasks, we need to relate information stored in a higher cortical area with the current incoming signal in a lower area.  

The diagram suggests strongly that the neurons in layers 2/3A and layer 6 are places where transformer-like algorithms can be implemented. Although many pyramidal cells in middle layers have long apical dendrites connecting the soma to layer 1 synapses at the end of feedback pathways, it is hard to see how the sparse signaling along the apical dendrite can allow very much integration of top-down and bottom-up data. But layer 2 pyramidal cells as well as multipolar layer 6 neurons have much more compact dendritic arbors and might do this.  Layer 6 feedback is perhaps the strongest candidate as this is less focused, more diffuse than layer 1 feedback (see Marcus et al). I strongly believe that some such mechanism must be used in mammalian cortex and that this is an exciting area for future research.

\section{Scaling}
If there is one thing human society and human economy teaches you, it is that scaling up any enterprise, any organization by a large factor requires many many adjustments, even radical re-organization. Most things don't just scale up easily. So how is it that the cerebral cortex of mammalian brains scales from a mouse with about 13 million cortical neurons to a human with about 16 billion cortical neurons, more than 3 orders of magnitude, with hardly any changes at all? The architecture of mammalian brains is totally different from those of birds and reptiles, so different that there are no universally agreed homologies between them. The mammalian neocortex just appeared from nowhere, apparently in its full blown form, same pyramidal cells, same 6 layers, same basic areas and same links to thalamus, etc. But once formed, almost all mammalian cerebral cortices seem essentially identical {\it except for size}. OK, the human brain has a uniquely large prefrontal lobe but this requires no major rewiring as well as a small set of peculiar ``von Economo cells'' and whale brains are an exception, simplifying their layer organization. But whatever algorithm makes mice smart seems to be the same thing that works for us humans too. 

A very simple observation, but one that I think is fundamental, is that present day AI, in both its functioning and its training, seems to have the same remarkable resilience to scaling. I like to demonstrate in my lectures the way neural nets work with a ``Mickey Mouse" example of a neural net with 12 weights that learns nearly perfectly in front of the live audience to discriminate points in the plane inside a circle from those outside, using simple gradient descent. OpenAI's most recent language program GPT-3 is based on the same ideas as BERT but has 175 {\it billion} weights and is trained by the same old gradient descent. Who would have expected that such a scaling was possible? The fact that simple minded gradient descent continues to work is astonishing. Yes, there are a few tricks like dropout, pre-training, etc. and OpenAI and Google have the best programmers tuning it up but it is basically still just gradient descent on very similar architectures.

\section{What is missing?}
Although on some problems with some measures, the so-called ``leader-board" shows AI programs approaching or even surpassing human skills, there are many ways in which they still fall far short of human skills. For example, GPT-3 when asked how many eyes your foot has, said your foot has two eyes. I guess they didn't train it on the classic folk song ``Dem dry bones'' or it might have had a little better anatomical knowledge under its ``belt''. 

\subsection{Vision problems}
More significantly, the idea of transformers hasn't made any impressive headway in computer vision where the central problem is segmenting and then identifying objects in possibly cluttered images. The google team have attacked vision with transformers, calling this self-attention\footnote{arXiv 1906.05909v1 and 1904.09925}. Their approach is to use the same architecture as the linguistic programs, but replace words by pixels, sentences by images. Self-attention seeks useful attention links from a vector of filter responses at one pixel to the same vector at some other pixel. This means transformers may need to link any pair of pixels, a huge challenge even for GPUs. Indeed a fundamental issue with all vision computations is handling the large size of data of a single image: any recognizable image needs a lot of pixels.  I have argued that because of the size of image data, animal cortices can only afford to have one set of neurons that keep full resolution. In other words, V1 must be the only high resolution buffer in which to do things that need accuracy (like comparing the proportions of two faces). Be this as it may, one should note that both the programs using transformers and conventional neural nets are already doing tremendously better than pre-2010 algorithms using hand designed filters instead using neural net learned filters. For example, the 2017 CVPR paper\footnote{arXiv 1612.03144v2} built on a {\it Feature Pyramid Network} (without transformers) outperforms all hand engineered programs on the so-called COCO benchmark and does combine all pyramid levels in one master representation.

My sense is that understanding static 2D images is a really tough skill to master. Dogs only rarely recognize the content of a photo, e.g. most dogs don't recognize photos of their masters, but they can crash through the cluttered woods at top speed. It's important to realize that human babies as well as dogs learn vision in a moving world (and also making use of the tectum, the reptilian brain stem visual structure). When either you or the perceived object moves, objects at different distances shift relative to each other and this makes it easy to separate figure and background. Further motion reveals their 3D shape. I suggest that transformers will solve vision problems better when trained on movies or from robots moving around, equipped with cameras. More data makes the task easier. Actually, babies start off in the cradle learning hand-eye coordination, using both external and internally generated motion. And they have stereo vision as well which amounts to seeing everything from two places, separated by a small movement. Similarly, the challenge of driving autonomous vehicles should be learnable by transformers and I'm sure this is being implemented somewhere even now. 

\subsection{General AI}
To analyze the next steps towards ``general AI'', let's consider the following model for the child's acquiring basic knowledge of the world around it. Starting with raw sensory input, the infant sees/hears/feels many confusing patterns, ``one great blooming, buzzing confusion'' as William James famously put it. But it soon recognizes simple recurring patterns. And then it sees patterns among the patterns, co-occurences, and learns to recognize larger more complex patterns. This leads to a tree in which various bits that have been linked are {\it reified} into a higher level concept. As it goes on, the resulting tree is very much like the parse trees in conventional grammar. Each new step results in learning what my colleague Stuart Geman calls ``reusable parts". It frequently happens that the pattern found in one context also occurs in a second quite different context. It is well established that in language acquisition, there are definite steps when the child acquires a new rule or concept and suddenly is able to apply it to new situations. This can be syntactical like seeing that most English verbs have a past tense formed by adding ``ed'' (love/loved) in contrast to a few very common exceptions (``see/saw''). A new word may be learned after only hearing it spoken {\it once}. Or it may be discovering a semantic class along with the word for the class, e.g.\ ``car". This process of growing your cognitive framework, often in discrete steps, continues your whole life.  Human brains do not even get fully connected until adolescence when the long distance axons that connect the most distant cortical areas are fully activated ({\it myelinated} is the technical term).

Much of this learning is already being done with neural nets. Really complex neural nets are being trained in stages. They may start with a net trained to answer simple low level questions about the data. Then layers are added that use the representations formed in the first net and are trained with more complicated questions. But suppose things computed in the higher layers suggest a modification to activity in the lower layers? In animal cortex, there is always feedback from the higher areas to which the lower areas project.  This suggests that a new kind of transformer is needed for this, something with queries and values in the original net and keys in the new higher layers. This creates circular computations and raises an issue of timing. However, this is a mechanism known to occur in the brain.

Another example is a robot learning hand-eye coordination. In humans, the infant connects efferent muscle signal patterns with afferent retinal stimulus, but this is a complex relationship and needs to be learned in order to coordinate activity in the corresponding visual and motor parts of the cortex. The robot may have a pretrained visual program and a pretrained motion program but now it needs to join them together with transformers that pick out aspects of each representation that the other needs to use. It needs to learn what muscle commands lead to what visual stimulus, and more, to merge the representations of space both nets have formed. 

In general, {\it distinct neural nets need some way to merge}, to train a larger net containing them both. As in the hand-eye situation, there may well be that some concepts  implicit in the distributed representations of both neural nets, but how would the nets ``know'' that they have hit on the same reusable idea?  Connecting two neural nets should certainly not need starting from scratch and relearning each set of weights. One needs instead to add new layers and transformers to create a larger net on top of the two others. I think this is an ideal task for a second generation of transformers, layered on top of the two pre-trained nets. The queries are in net \#1, the keys in net \#2 (or vice versa) and the training involves tasks where both nets need to work together. In terms of graphs, a parse tree or AND/OR graph should be present implicitly in the representations of the two nets and these new transformers should find the common nodes leading to the creation of a larger, but still implicit, graph for the merged net. 

The issue of feedback is central in the task of comparing memories with current stimulus. At every instant, you are usually experiencing a new configuration of events related to various old events and you merge memory traces as much as possible with the new sensory input and new situation, a process that appears to be mediated by feedback between neurons as we discussed above. For example, everyone has an inventory of known faces, e.g. the faces of your family, friends and co-workers. They are likely stored in the fusiform face area (FFA) or adjacent areas of inferior temporal cortex. When you see them again, you must perform some sort of matching before you can say you recognize them. As shapes, sizes, relative positions are involved, my own belief is that V1 must play a role via feedback, all the way from FFA or IT.  But in all cases, the memories will not match the new stimulus exactly: there will always be changes, they will not be exact repeats. You must notice the changes in order to understand best what's happening now. This role of feedback -- noticing the differences -- was central in my papers referred to above. Is this needed and, if so, will transformers be needed for this? 

All of this suggests that to reach general AI, neural nets will need to have something like memory in higher levels and feedback to lower levels, to be more modular and to have structures specific for both feedforward and feedback data exchange. So far, neural nets have been only a little modular. BERT has two pieces, the encoder and the decoder, and  recent segmentation algorithms have more, some even looking a bit like analogs of the distinct mammalian visual areas V1, V2, V4. How many would be needed if general AI is achieved?, a big question.  Finally, cortical architecture has a very specific architecture with the hippocampus acting like the highest layer, storing current memories for variable periods but eventually downloading some of them into appropriate cortical areas, forgetting many others. Should AI's imitate this if they seek human level skills? 

\subsection{The Whorfian hypothesis}
I want to go back to {\it word2vec} where the idea was that high dimensional vectors are better carriers of linguistic data than the discrete tokens called words. There is one school of thought that asserts the opposite: that words are what has enabled humans to think so well and that, as a result, the way you conceptualize something mimics how your language expresses it. This is the so-called {\it Whorfian Hypothesis}, named after the linguist and engineer Benjamin Whorf who developed this idea together with Edmund Sapir. To some extent, this feels right, that it expresses well the content of consciousness. And yet, often words just come out of your mouth unconsciously, without any reflection, any sense of your having had a choice. Your consciousness then looks like a supervisor watching what emerges from the unseen machines grinding away below.  This is the model Stanislav Dehaene propounds in his book {\it Consciousness and the Brain}. In other words, we can understand thought either as manipulating word tokens  \`a la Whorf, or, alternatively, think of the words as a gloss your consciousness puts on the output of a vast set of firing neurons, a sort of executive summary, \`a la Dehaene.  From a computational perspective, this is simply the choice between discrete token-based representations and distributed real vector representations. 

My own belief is that distributed representations are here to stay. I see no reason why we need single neurons or single neural net units that learn to respond to unique features of an ongoing thought process. Yes, we need outputs made of discrete signals. In brains, it seems that Broca's area processes a distributed representation of a thought into a grammatical utterance made from a sequence of words; and Google's BERT has a decoder half that outputs a sentence, retrieving the word tokens, so-to-speak, at the last minute. This is all about as far from Chomsky's Universal Grammar and from the Whorf-Sapir theory as it could be. It asserts that we know the grammar of our mother tongue not by rules but by endless experiences of its usage and nuances and by playing the game of sometimes using precise rules, sometimes ignoring them\footnote{My favorite example, given me by Jean Berko Gleason, is ``This dress zips up the back'', making the dress into an active agent, when the passive voice was obviously called for.}. It has emerged in the last few decades how much cortical activity is unconscious, how little makes its way into consciousness. Maybe what we are conscious of is the output of a decoder, like that in BERT, and is more token-like while the unconscious stuff are all embodied in distributed representations.

What is even more striking, disconcerting in fact, is that, if this is true, then all future AI machines will be hard to impossible to understand, to know why ``it'', the machine, has concluded something.  Indeed, we are truly living in a ``Brave new world" with wonders aplenty. 

\end{document}